\documentclass[11pt,a4paper]{article}
\usepackage[hyperref]{naaclhlt2018}
\usepackage{times}
\usepackage{latexsym}
\usepackage{amsmath}
\usepackage{multirow}
\usepackage{url}
\usepackage[T1]{fontenc}    
\usepackage{hyperref}       
\usepackage{booktabs}       
\usepackage{amsfonts}       
\usepackage{nicefrac}       
\usepackage{microtype}      
\usepackage{mathtools}
\usepackage{graphicx,color,xcolor}
\usepackage{subfigure}
\usepackage{enumitem} 
\usepackage{bm}
\usepackage{algorithm}
\usepackage[noend]{algpseudocode}
\usepackage{todonotes}
\usepackage{tablefootnote}
\usepackage{relsize}
\usepackage{footnote}
\usepackage{bbding}
\usepackage{xcolor,colortbl}
\makesavenoteenv{tabular}
\makeatletter
\newcommand{\@emptybiblabel}[1]{}
\newcommand*\samethanks[1][\value{footnote}]{\footnotemark[#1]}
\makeatother

\aclfinalcopy 
\def\aclpaperid{446} 

\title{Universal Neural Machine Translation for \\Extremely Low Resource Languages}
\def \msr{$^\ddag$}
\def \hku{$^\dagger$}
\def \goo{$^\square$}
\author{Jiatao Gu\hku\thanks{~~This work was done while the authors at Microsoft.}~~~~ Hany Hassan\msr ~~~~ Jacob Devlin\goo\samethanks[1] ~~~~ Victor O.K. Li\hku 
\\
{\hku {The University of Hong Kong}} ~~~~~~~~~~ {\msr{Microsoft Research}}\\
{ \tt \{jiataogu, vli\}@eee.hku.hk } ~~  \tt hanyh@microsoft.com\\
{ \goo{Google Research}} \\
{ \tt jacobdevlin@google.com}\\
}

\date{}

\begin{document}
\maketitle
\begin{abstract}
 In this paper, we propose a new universal machine translation approach focusing on languages with a limited amount of parallel data.  Our proposed approach utilizes a transfer-learning approach to share lexical and sentence level representations across multiple source languages into one target language. The lexical part is shared through a  Universal Lexical Representation to support multi-lingual word-level sharing. The sentence-level sharing is represented by a model of experts from all source languages that share the source encoders with all other languages. This enables the low-resource language to utilize the lexical and sentence representations of the higher resource languages. 
Our approach is able to achieve 23 BLEU on Romanian-English WMT2016 using a tiny parallel corpus of 6k sentences, compared to the 18 BLEU of strong baseline system which uses multi-lingual training and back-translation. Furthermore, we show that the proposed approach can achieve almost 20 BLEU on the same dataset through fine-tuning  a pre-trained multi-lingual system in a zero-shot setting.

\end{abstract}

\section{Introduction}
Neural Machine Translation (NMT)~\cite{bahdanau2014neural} has achieved remarkable  translation quality in various  on-line large-scale systems~\cite{wu2016google,devlin:2017:EMNLP2017} as well as achieving  state-of-the-art results on Chinese-English  translation ~\cite{hassan-hp}. With such large systems, NMT showed that it can scale up to immense  amounts of parallel data in the order of tens of millions of sentences. However, such data is not widely available for all language pairs and  domains. In this paper, we propose a novel universal multi-lingual NMT approach   focusing mainly on low resource languages to overcome the  limitations of NMT and leverage the capabilities of multi-lingual NMT  in such scenarios.

Our approach utilizes multi-lingual neural translation system to share lexical and sentence level representations across multiple source languages into one target language. In this setup, some of the source languages may be of extremely limited  or even zero data.  The lexical sharing is represented by a  universal word-level representation where various words from all source languages  share the same underlaying representation. The sharing module utilizes monolingual embeddings along with seed parallel data from all languages to build the universal representation. The sentence-level sharing is represented by a model of language experts which  enables low-resource  languages to  utilize the sentence representation of the higher resource languages.  This allows the system to translate from any language even with tiny amount of parallel resources.  

We evaluate the proposed approach on 3 different  languages with tiny or even zero parallel data.
We show that for the simulated ``zero-resource" settings, our model can consistently outperform a strong multi-lingual NMT baseline with a tiny amount of parallel sentence pairs.

\section{Motivation}
Neural Machine Translation
(NMT)~\cite{bahdanau2014neural,sutskever2014sequence}  is based on Sequence-to-Sequence encoder-decoder model along with an attention mechanism to enable better  handling of  longer sentences \cite{bahdanau2014neural}. Attentional sequence-to-sequence models are modeling the log conditional probability of the translation $Y$ given an input sequence $X$.  
In general, the NMT system $\theta$ consists of two components: an encoder $\theta_e$ which transforms the input sequence into an array of continuous representations,
and a decoder $\theta_d$ that dynamically reads the encoder's output with an attention mechanism and predicts the distribution of each target word. 
Generally, $\theta$ is trained to maximize the likelihood on a training set consisting of $N$ parallel sentences: 
\begin{equation}
	\begin{split}
	&\mathcal{L}\left(\theta\right)=\frac{1}{N}\sum_{n=1}^N\log p\left(Y^{(n)}|X^{(n)}; \theta\right) \\
    &=\frac{1}{N}\sum_{n=1}^N\sum_{t=1}^T\log p\left(y_t^{(n)}|y_{1:t-1}^{(n)}, f^{\text{att}}_t(h^{(n)}_{1:T_s})\right)
	\end{split}
	 \label{eq.loss} 
\end{equation}
where at each step, $f^{\text{att}}_t$ builds the attention mechanism over the encoder's output $h_{1: T_s}$.
More precisely, let the vocabulary size of source words as $V$
\begin{equation}
h_{1: T_s} = f^{\text{ext}}\left[e_{x_1},..., e_{x_{T_s}} \right], \ \ \ e_x = E^I(x)
\label{eq.encoder}
\end{equation}
where $E^I \in \mathbb{R}^{V \times d}$ is a look-up table of source embeddings, assigning each individual word a unique embedding vector; $f^{\text{ext}}$ is a sentence-level feature extractor and is usually implemented by a multi-layer bidirectional RNN~\cite{bahdanau2014neural,wu2016google}, recent efforts also achieved the state-of-the-art using non-recurrence $f^{\text{ext}}$, e.g. ConvS2S~\cite{gehring2017convolutional}  and Transformer~\cite{vaswani2017attention}.
\begin{figure}[t]
	\includegraphics[width=\linewidth]{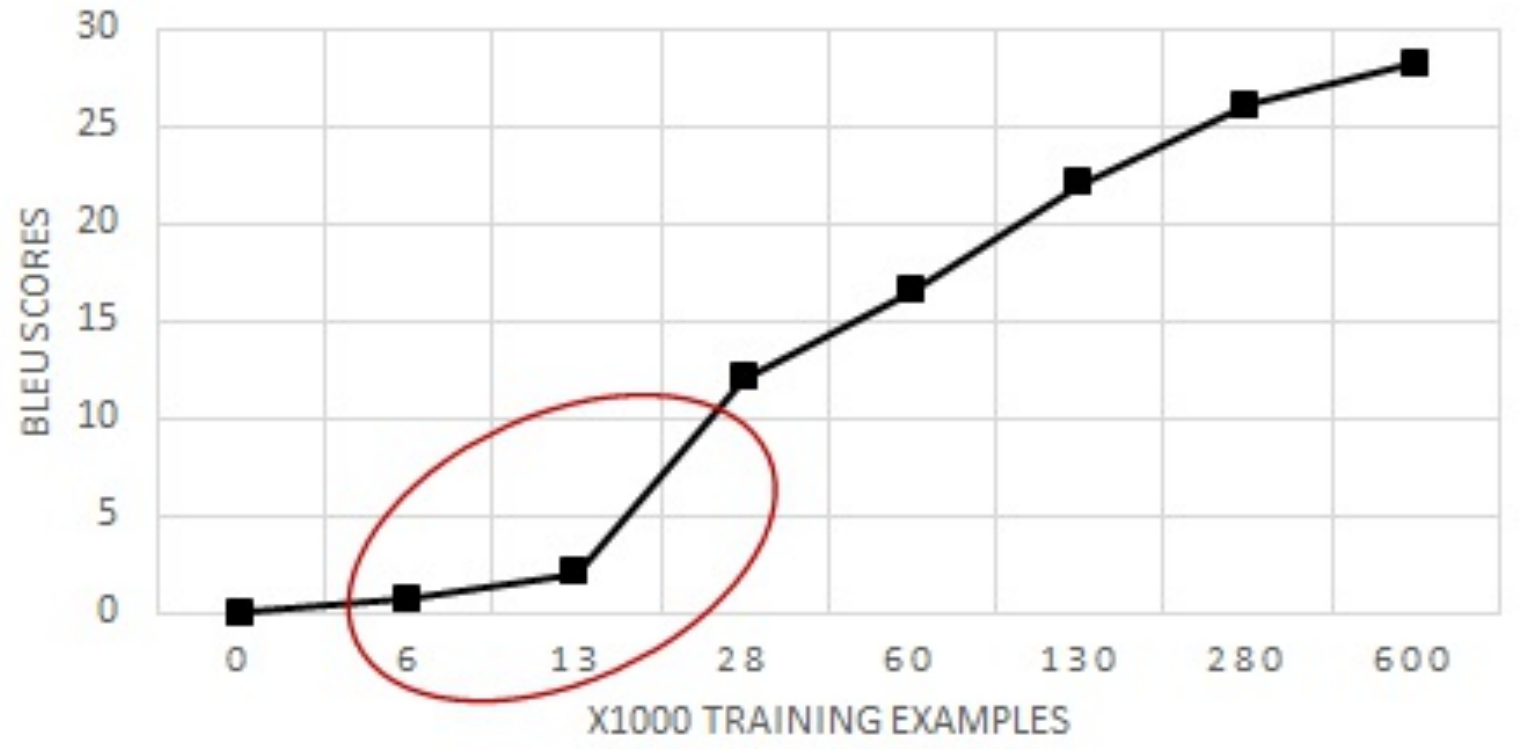}
\caption{\label{fig.data_size} BLEU scores reported on the test set for Ro-En. The amount of training data effects the translation performance dramatically using a single NMT model.}\vspace{-10pt}	
 \end{figure}

\paragraph{Extremely Low-Resource NMT} Both $\theta_e$ and $\theta_d$ should be trained to converge using parallel training examples. However, the performance is highly correlated to the amount of training data.  As shown in Figure.~\ref{fig.data_size}, the system cannot achieve reasonable translation quality when the number of the parallel examples is extremely small ($N \approx 13k$ sentences,  or not available at all $N =0$).

\paragraph{Multi-lingual NMT}\newcite{lee2016fully} and \newcite{johnson2016google} have shown that NMT is quite efficient for  multilingual machine translation. Assuming the translation from $K$ source languages into one target language, a  system is trained with maximum likelihood on the mixed parallel pairs $\{X^{(n, k)}, Y^{(n, k)}\}_{k=1 ... K}^{n=1 ... N_k}$, that is
\begin{equation}
	\mathcal{L}\left(\theta\right)=\frac{1}{N}\sum_{k=1}^{K}\sum_{n=1}^{N_k}\log p\left(Y^{(n, k)}|X^{(n, k)}; \theta\right)
\end{equation}
where $N=\sum_{k=1}^K N_k$. As the input layer, the system assumes a multilingual vocabulary which is usually the union of all  source language vocabularies with a total size as $V=\sum_{k=1}^K V_k$. In practice, it is essential to shuffle the multilingual sentence pairs into mini-batches so that different languages can be trained equally.
Multi-lingual NMT is quite appealing for low-resource languages; several papers highlighted  the characteristic that make it a good fit for that  such as  \newcite{lee2016fully}, \newcite{johnson2016google}, \newcite{zoph2016transfer} and \newcite{firat2016multi}. Multi-lingual NMT utilizes the training examples of multiple languages to regularize the models  avoiding over-fitting to the limited data of the smaller languages. Moreover, the model transfers the translation knowledge from high-resource languages to low-resource ones. Finlay, the decoder part of the model is sufficiently trained  since it shares  multilingual examples from all languages.

\subsection{Challenges}
Despite the success of training multi-lingual NMT systems; there are a couple of challenges to leverage them for zero-resource languages:

\paragraph{Lexical-level Sharing} Conventionally, a multi-lingual NMT model has a vocabulary that represents the union of the vocabularies of all source languages. Therefore, the multi-lingual words do not practically share the same embedding space since each word has its own representation. This does not pose a problem for languages  with sufficiently large amount of  data, yet it is a major limitation for extremely low resource languages since most of the vocabulary items will not have enough, if any, training examples to get a reliably trained models.

A possible solution is to share the surface form of  all source languages through sharing sub-units such as subwords ~\cite{sennrich2015neural} or characters~\cite{kim2016character,luong2016achieving,lee2016fully}.  
However, for an arbitrary low-resource language we cannot assume significant overlap in the lexical surface forms compared to the high-resource languages. The low-resource language may not even share the same character set as any high-resource language. It is crucial to create a shared semantic representation across all languages that does not rely on surface form overlap.

\begin{figure*}[t]
	\centering
	\includegraphics[width=0.98\linewidth]{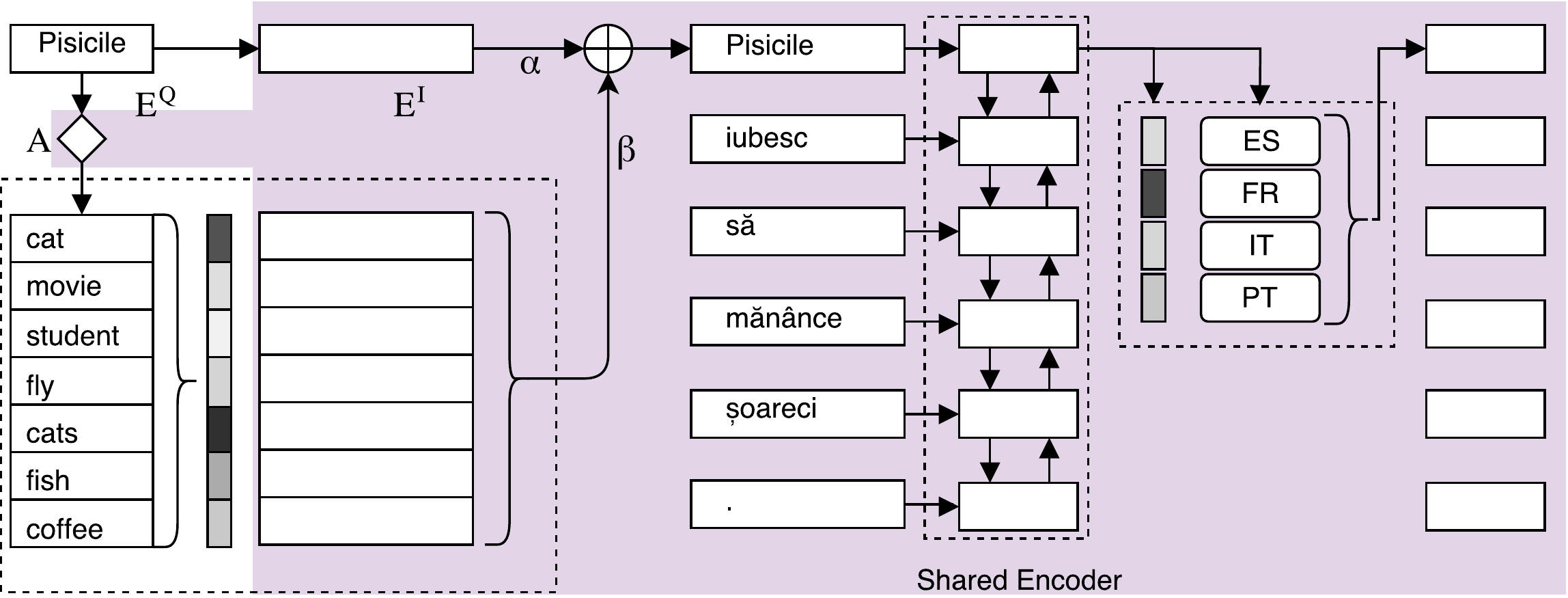}\vspace{-15pt}
      \caption{\label{fig.model} An illustration of the proposed architecture of the ULR and MoLE. Shaded parts are trained within NMT model while unshaded parts are not changed during  training.}\vspace{-8pt}
\end{figure*}

\paragraph{Sentence-level Sharing} It is also crucial for low-resource languages to share source sentence representation with other similar languages. For example, if a language shares  syntactic order with another language it should be feasible for the low-resource language to share such representation with another high recourse language. It is also important to utilize monolingual data to learn such representation since the low or zero resource language may  have monolingual resources only.


\section{Universal Neural Machine Translation}
We propose a Universal NMT system that is focused on the scenario where minimal parallel sentences are available. 
As shown in Fig.~\ref{fig.model}, we introduce two components to extend the conventional multi-lingual NMT system \cite{johnson2016google}: Universal Lexical Representation (ULR) and Mixture of Language Experts (MoLE) to enable both word-level and sentence-level sharing, respectively.

\subsection{Universal Lexical Representation (ULR)}
\label{sec.unilex}

As we highlighted above, it is not straightforward to have a universal representation  for all languages. One potential approach is to use a shared source vocabulary, but this is not adequate since it assumes significant  surface-form overlap in order being able to generalize between high-resource and low-resource languages. Alternatively, we could train monolingual embeddings in a shared space and use these as the input to our MT system. However, since these embeddings are trained on a monolingual objective, they will not be optimal for an NMT objective. If we simply allow them to change during NMT training, then this will not generalize to the low-resource language where many of the words are unseen in the parallel data.
Therefore, our goal is to create a shared embedding space which (a) is trained towards NMT rather than a monolingual objective, (b) is not based on lexical surface forms, and (c) will generalize from the high-resource languages to the low-resource language. 

We propose a novel representation for multi-lingual embedding where each word from any language is represented as a probabilistic mixture of universal-space word embeddings. In this way, semantically similar words from different languages will naturally have similar representations. Our  method achieves this  utilizing a discrete (but probabilistic) ``universal token space'', and then learning the embedding matrix for these universal tokens directly in our NMT training.

\paragraph{Lexicon Mapping to the Universal Token Space}
We first define a discrete universal token set of size $M$ into which all source languages will be projected. In principle, this could correspond to any human or symbolic language, but all experiments here use English as the basis for the universal token space. As shown in Figure \ref{fig.model}, we have multiple embedding representations. $E^Q$ is language-specific embedding trained on  monolingual data and $E^K$ is universal tokens embedding. The matrices $E^K$ and $E^Q$ are created beforehand and are not trainable during NMT training.  $E^U$ is the embedding matrix for these universal tokens which is learned during  our NMT training.  It is worth noting that shaded parts in Figure\ref{fig.model} are trainable during  NMT training process.

Therefore, each source word  $e_x$ is represented as a mixture of universal tokens $M$ of $E^U$.

\begin{equation}
	e_x = \sum_{i=1}^M E^U\left(u_i\right) \cdot q(u_i|x)
    \label{eq.universal_embed}
\end{equation}
where $E^U$ is an NMT embedding matrix, which is learned during NMT training.

The mapping $q$ projects the multilingual words into the universal space based on their semantic similarity. That is, $q(u|x)$ is a distribution based on the distance $D_s(u, x)$ between $u$ and $x$ as:
\begin{equation}
 q(u_i|x) = \frac{e^{D(u_i, x) / \tau}}{\sum_{u_j} e^{D(u_j, x) / \tau}}
 \label{eq.q_softmax}
\end{equation}
where $\tau$ is a temperature and $D(u_i, x)$ is a scalar score which represents the similarity between source word $x$ and universal token $u_i$:
\begin{equation}
\label{eq.ds}
	D(u, x) =  E^K(u)\cdot A\cdot E^Q(x)^T
\end{equation}
where $E^K(u)$ is the ``key'' embedding of word $u$, $E^Q(x)$ is the ``query'' embedding of source word $x$.  The transformation matrix $A$, which is initialized to the identity matrix, is learned during NMT training and shared across all languages. 

This is a key-value representation, where the queries  are the monolingual language-specific embedding, the keys are the universal tokens embeddings and the values are a probabilistic distribution over the universal NMT embeddings. 
This can  represent  unlimited  multi-lingual vocabulary  that has  never been observed in the parallel training data.   
It is worth noting  that the trainable transformation matrix $A$ is added  to the query matching mechanism  with the main purpose to tune the similarity scores towards the translation task. 
$A$ is shared across all languages and optimized discriminatively during NMT training such that the system can fine-tune the similarity score $q()$ to be optimal for NMT.



\paragraph{Shared Monolingual Embeddings}
In general, we create one $E^Q$ matrix per source language, as well as a single $E^K$ matrix in our universal token language. For Equation~\ref{eq.ds} to make sense and generalize across language pairs, all of these embedding matrices must live in a similar semantic space. To do this, we first train off-the-shelf monolingual word embeddings in each language, and then learn one projection matrix per source language which maps the original monolingual embeddings into $E^K$ space.
Typically, we need a list of \textit{source word - universal token} pairs (seeds $S_k$) to train the projection matrix for language $k$. Since vectors are normalized, learning the optimal projection is equivalent to finding an orthogonal transformation $O_k$ that makes the projected word vectors as close as to its corresponded universal tokens:
\begin{equation}
  \begin{array}{l}  
         \max\limits_{O_k}\mathlarger{\sum}\limits_{(\tilde{x}, \tilde{y})\in S_k}  \left(E^{Q_k}(\tilde{x})\cdot O_k\right)\cdot E^K(\tilde{y})^T \\  
         \text{s.t.       } O_k^TO_k = I, \ \ \ k=1, ..., K
 \end{array}  
\end{equation}
which can be solved by SVD decomposition based on the seeds~\cite{smith2017offline}. In this paper, we chose to use a short list of seeds from automatic word-alignment of parallel sentences  to learn the projection. However, recent efforts~\cite{Artetxe2017LearningBW,Conneau2017WordTW}   also showed that it is possible to learn the transformation without any seeds, which makes it feasible  for our  proposed method to be utilized in purely zero parallel resource cases.

It is worth noting that  $O_k$ is a language-specific matrix which maps the monolingual embeddings of each source language into a similar semantic space as the universal token language.

\paragraph{Interpolated Embeddings}
Certain lexical categories (e.g. function words) are poorly captured by Equation~\ref{eq.universal_embed}. Luckily, function words often have very high frequency, and can be estimated robustly from even a tiny amount of data. This motivates an interpolated $e_x$ where embeddings for very frequent words are optimized directly and  not through the universal tokens:
\begin{equation}
	\alpha(x) E^I(x) + \beta(x) \sum_{i=1}^M E^U\left(u_i\right) \cdot q(u_i|x)
\end{equation}
Where $E^I(x)$ is a language-specific embedding of word $x$ which is optimized during NMT training. In general, we set $\alpha(x)$ to 1.0 for the top $k$ most frequent words in each language, and 0.0 otherwise, where $k$ is set to 500 in this work. It is worth noting that we do not use an absolute frequency cutoff because this would cause a mismatch between high-resource and low-resource languages, which we want to avoid. We keep $\beta(x)$ fixed to 1.0.

\paragraph{An Example} To give a concrete example, imagine that our target language is English (En), our high-resource auxiliary source languages are Spanish (Es) and French (Fr), and our low-resource source language is Romanian (Ro). En is also used for the universal token set. We assume to have 10M+ parallel Es-En and Fr-En, and a few thousand in Ro-En. We also have millions of monolingual sentences in each language.

We first train word2vec embeddings on monolingual corpora from each of the four languages. We next align the Es-En, Fr-En, and Ro-En parallel corpora and extract a seed dictionary of a few hundred words per language, e.g., ${\tt gato} \rightarrow {\tt cat}$,  ${\tt chien} \rightarrow {\tt dog}$. We then learn three matrices $O_1, O_2, O_3$ to project the Es, Fr and Ro embeddings ($E^{Q_1}, E^{Q_2}, E^{Q_3}$), into En ($E^K$) based on these seed dictionaries. At this point, Equation~\ref{eq.q_softmax} should produce \textit{reasonable} alignments between the source languages and En, e.g., $q(\tt{horse}|\tt{magar}) = 0.5$, $q(\tt{donkey}|\tt{magar}) = 0.3$, $q(\tt{cow}|\tt{magar}) = 0.2$, where {\tt magar} is the Ro word for {\tt donkey}. 

\subsection{Mixture of Language Experts (MoLE)}
\label{sec.moe}
As we paved the road for having a universal embedding representation; it is crucial to have a  language-sensitive module for the encoder that would help in modeling various  language structures which may  vary between different languages. 
We propose a Mixture of Language Experts (MoLE) to model the sentence-level universal encoder. As shown in Fig.~\ref{fig.model}, 
an additional module of mixture of experts is used after the last layer of the encoder. Similar to \cite{shazeer2017outrageously}, we have a set of expert networks and a gating network  to control the weight of each expert. More precisely, we have a set of expert networks as $f_1(h), ..., f_{K}(h)$ where for each expert, a two-layer feed-forward network which reads the output hidden states $h$ of the encoder is utilized. The output of the MoLE module $h'$ will be a weighted sum of these experts to replace the encoder's representation:
\begin{equation}
h'=\sum_{k=1}^K f_k(h)\cdot \text{softmax}(g(h))_k,
\end{equation}
where an one-layer feed-forward network $g(h)$ is used as a gate to compute scores for all the experts.

In our case, we create one expert per auxiliary language. In other words, we train to only use expert $f_i$ when training on a parallel sentence from auxiliary language $i$. Assume the language $1 ... K-1$ are the auxiliary languages. That is, we have a multi-task objective as:
\begin{equation}
\begin{split}
\mathcal{L}^{\text{gate}} = \sum_{k=1}^{K-1}\sum_{n=1}^{N_k}\log \left[\text{softmax}\left(g(h)\right)_{k}\right]
\end{split}
\end{equation}

We do not update the MoLE module for training on a sentence from the low-resource language. Intuitively, this allows us to represent each token in the low-resource language as a context-dependent mixture of the auxiliary language experts.


\section{Experiments}
\label{sec.exps}
We extensively study the effectiveness of the proposed methods by evaluating on three ``almost-zero-resource'' language pairs with variant auxiliary languages. The vanilla single-source NMT and the multi-lingual NMT models are used as baselines.
\subsection{Settings}

\paragraph{Dataset} We empirically evaluate the proposed Universal NMT system on $3$ languages -- Romanian (Ro) / Latvian (Lv) / Korean (Ko)  -- translating to English (En) in near zero-resource settings. To achieve this, single or multiple auxiliary languages from Czech (Cs), German (De), Greek (El), Spanish (Es), Finnish (Fi), French (Fr),  Italian (It), Portuguese (Pt) and Russian (Ru) are jointly trained. The detailed statistics and sources of the available parallel resource can be found in Table~\ref{table.data}, where we further down-sample the corpora for the targeted languages to simulate zero-resource. 

\begin{savenotes}
\begin{table*}[t]
\centering
\small
\begin{tabular}{p{0.9cm}||*{13}{p{0.5cm}}}
\hline
& \multicolumn{3}{c|}{Zero-Resource Translation} & \multicolumn{9}{c}{Auxiliary High-Resource Translation} \\ \hline
source &  
\multicolumn{1}{c|}{Ro} & \multicolumn{1}{c|}{Ko} & \multicolumn{1}{c|}{Lv} & 
\multicolumn{1}{c|}{Cs} & \multicolumn{1}{c|}{De} & \multicolumn{1}{c|}{El}  & \multicolumn{1}{c|}{Es}  & \multicolumn{1}{c|}{Fi} & 
\multicolumn{1}{c|}{Fr} & \multicolumn{1}{c|}{It} & \multicolumn{1}{c|}{Pt} &  
\multicolumn{1}{c}{Ru}  \\ \hline
corpora & \multicolumn{1}{c|}{WMT16\footnote{http://www.statmt.org/wmt16/translation-task.html}} & \multicolumn{1}{c|}{KPD\footnote{https://sites.google.com/site/koreanparalleldata/}} &  \multicolumn{9}{c|}{Europarl v8\footnote{http://www.statmt.org/europarl/}} & \multicolumn{1}{c}{UN \footnote{http://opus.lingfil.uu.se/MultiUN.php (subset)}} \\ \hline
size 
& \multicolumn{1}{c|}{612k} & \multicolumn{1}{c|}{97k} & \multicolumn{1}{c|}{638k} &  \multicolumn{1}{c|}{645k} & \multicolumn{1}{c|}{1.91m} 
& \multicolumn{1}{c|}{1.23m} & \multicolumn{1}{c|}{1.96m} & \multicolumn{1}{c|}{1.92m} & \multicolumn{1}{c|}{2.00m} & \multicolumn{1}{c|}{1.90m} 
& \multicolumn{1}{c|}{1.96m} & \multicolumn{1}{c}{11.7m}  \\ \hline
subset & \multicolumn{1}{c|}{0/6k/60k} & \multicolumn{1}{c|}{10k} & \multicolumn{1}{c|}{6k} 
&\multicolumn{8}{c}{/} &\multicolumn{1}{|c}{2.00m}\\ \hline
\end{tabular}
\vspace{-5pt}
\caption{\label{table.data}Statistics of the available parallel resource in our experiments. 	All the languages are translated to English.}
\end{table*}
\end{savenotes}

It also requires additional large amount of monolingual data to obtain the word embeddings for each language, where we use the latest Wikipedia dumps~\footnote{https://dumps.wikimedia.org/} for all the languages. Typically, the monolingual corpora are much larger than the parallel corpora. For validation and testing, the standard validation and testing sets are utilized for each targeted language.

\paragraph{Preprocessing}
All the data (parallel and monolingual) have been tokenized and segmented into subword symbols using byte-pair encoding (BPE)~\cite{sennrich2015neural}. We use sentences of length up to 50 subword symbols for all languages.  For each language, a maximum number of $40,000$ BPE operations are  learned and  applied to restrict the size of the vocabulary.  We concatenate the vocabularies of all source languages in the multilingual setting where special a ``language marker " have been appended to each word  so that there will be no embedding sharing on the surface form. Thus, we avoid sharing the representation  of words that have similar surface forms though with different meaning in various languages.

\paragraph{Architecture} We implement an attention-based neural machine translation model which consists of a one-layer bidirectional RNN encoder and a two-layer attention-based RNN decoder. All  RNNs have 512 LSTM units~\cite{hochreiter1997long}. Both the dimensions of the source and target embedding vectors are set to 512. The dimensionality of universal embeddings is also the same. For a fair comparison, the same architecture is also utilized for training both the vanilla and multilingual NMT systems. For multilingual experiments, $1\sim 5$ auxiliary languages are used.  When training with the universal tokens, the temperature $\tau$ (in Eq.~\ref{eq.ds}) is fixed to $0.05$ for all the experiments.

\paragraph{Learning}
All the models are trained to maximize the log-likelihood using Adam~\cite{kingma2014adam} optimizer for 1 million steps on the mixed dataset with a batch size of 128. The dropout rates for both the encoder and the decoder is set to 0.4. 
We have open-sourced an implementation of the proposed model.~\footnote{https://github.com/MultiPath/NA-NMT/tree/universal\_translation}

\subsection{Back-Translation}
We utilize back-translation (BT)~\cite{sennrich2016edinburgh} to encourage the model to use more information of the zero-resource languages. More concretely, we build the synthetic parallel corpus by translating on monolingual data\footnote{We used News Crawl provided by WMT16 for Ro-En.} with a trained translation system and use it to train a backward direction translation model. Once trained, the same operation can be used on the forward direction. 
Generally, BT is difficult to apply for zero resource setting since it requires a reasonably good translation system to generate good quality synthetic parallel data. Such a system may not be feasible with tiny or zero parallel data. However, it is possible to start with a trained multi-NMT model.

\subsection{Preliminary Experiments}
\paragraph{Training Monolingual Embeddings} We train the monolingual embeddings using \texttt{fastText}\footnote{https://github.com/facebookresearch/fastText}~\cite{bojanowski2016enriching}  over the Wikipedia corpora of all the languages.  The vectors are set to 300 dimensions, trained using the default setting of skip-gram . All the vectors are normalized to norm $1$.

\paragraph{Pre-projection} In this paper, the pre-projection requires initial word alignments (seeds) between words of each source language and the universal tokens.  More precisely, for the experiments of Ro/Ko/Lv-En, we use the target language (En) as the universal tokens;  \texttt{fast\_align}\footnote{https://github.com/clab/fast\_align} is used to automatically collect the aligned words between the source languages and English. 

\section{Results}

\begin{table}[t]
\centering
\begin{tabular}{l|c|rrr}
Src & Aux   & Multi & +ULR & + MoLE \\ \hline
\multirow{ 4}{*}{Ro}     
& Cs De El Fi & & 18.02 & 18.37   \\
& Cs De El Fr & & 19.48 &  19.52  \\
& De El Fi It & & 19.11 & 19.33   \\
& Es Fr It Pt & 14.83   & 20.01   &  \textbf{20.51} \\ \hline
\multirow{ 2}{*}{Lv}      
& Es Fr It Pt & 7.68     & 10.86     &  11.02 \\ 
& Es Fr It Pt Ru & 7.88     & 12.40     &  \textbf{13.16} \\  \hline
Ko    & Es Fr It Pt  & 2.45    & 5.49    & \textbf{6.14}  \\     
\end{tabular}
\caption{\label{table.bleu} Scores over variant source languages (6k sentences for Ro \& Lv, and 10k for Ko). ``Multi" means the Multi-lingual NMT baseline.}
\end{table}

We show our main results of multiple source languages to English with different auxiliary languages in Table~\ref{table.bleu}. To have a fair comparison, we use only 6k sentences corpus for both Ro and Lv with all the settings and 10k for Ko. It is obvious that applying both the universal tokens and mixture of experts modules  improve the overall translation quality for all the language pairs and the improvements are additive. 

 \begin{figure*}[t]
	\centering
    \begin{minipage}[t]{0.48\textwidth}
    \centering
    \includegraphics[width=\linewidth]{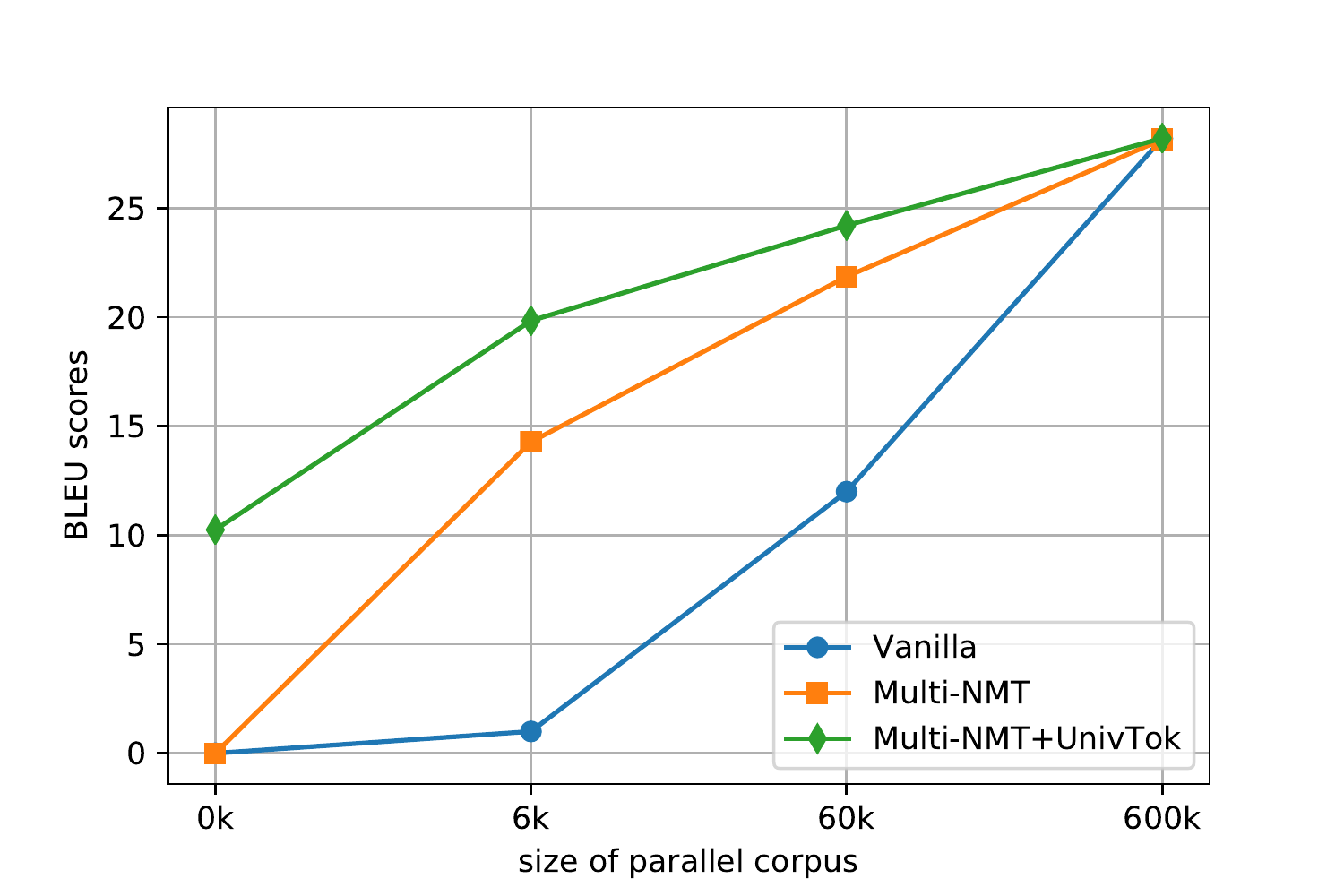}\hspace{-5pt}
    \caption{\label{fig.size}BLEU score vs corpus size}
    \end{minipage}
    \begin{minipage}[t]{0.48\textwidth}
    \centering
    \includegraphics[width=\linewidth]{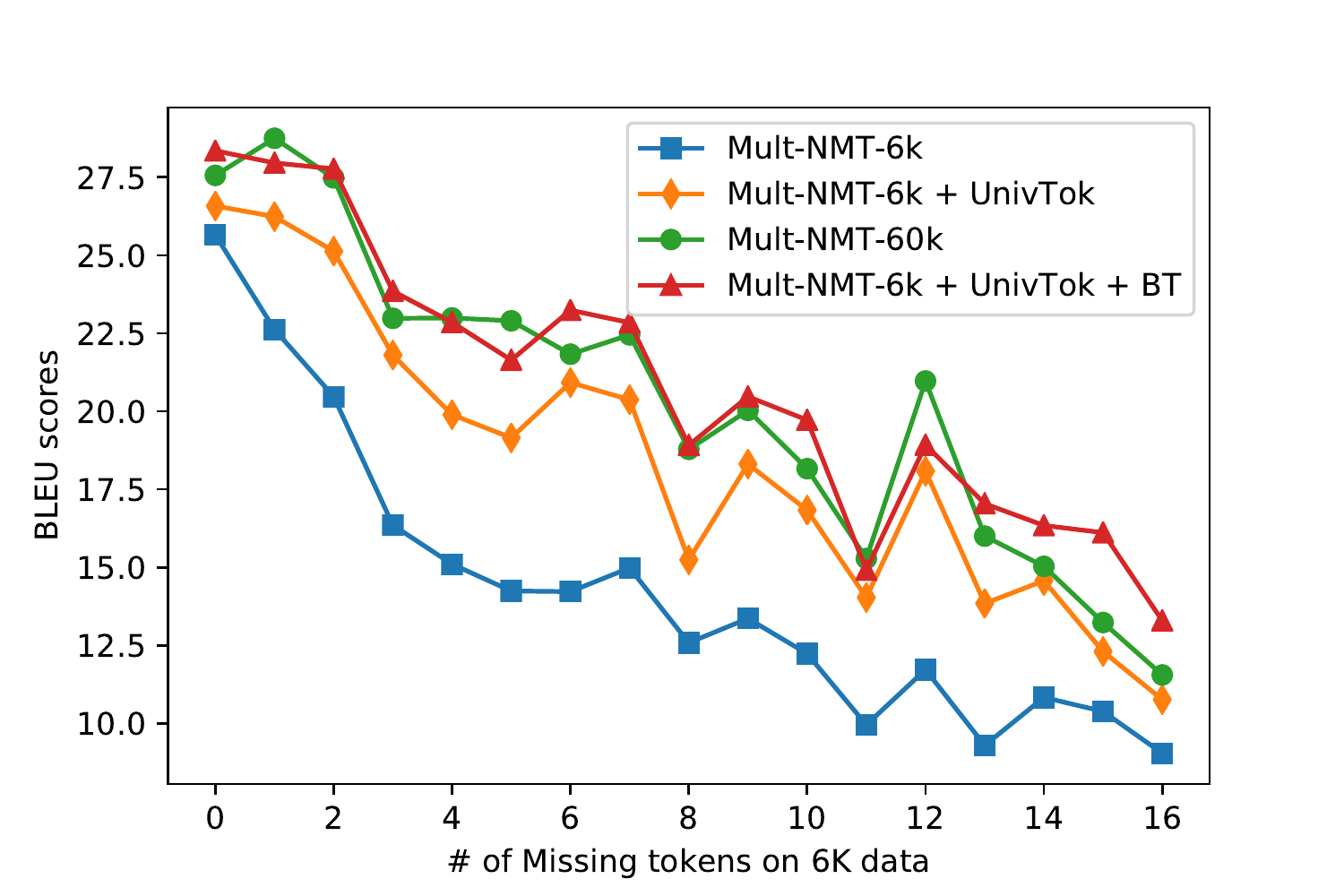}\hspace{-5pt}
    \caption{\label{fig.missing}BLEU score vs unknown tokens}
    \end{minipage}
\end{figure*}

To examine the influence  of auxiliary languages, we tested four sets of different combinations of auxiliary languages for Ro-En and two sets for Lv-En. It shows that Ro performs best when the auxiliary languages are all selected in the same family (Ro, Es, Fr, It and Pt are all from the Romance family of European languages) which makes sense as more knowledge can be shared across the same family. Similarly, for the experiment of Lv-En, improvements are also observed when adding Ru as additional auxiliary language as Lv and Ru share many similarities because of the geo-graphical influence even though they don't share the same alphabet. 

We also tested a set of Ko-En experiments to examine the generalization capability of our approach on  non-European languages while using languages of Romance family as auxiliary languages. Although the BLEU score is relatively low, the proposed methods can consistently help  translating less-related low-resource languages. It is more reasonable to have  similar languages as auxiliary languages.

\subsection{Ablation Study}
\begin{table}[t]
 	\centering
    \begin{tabular}{l|r}
    Models                   & BLEU  \\ \hline
    Vanilla                        & 1.21   \\
    Multi-NMT                 & 14.94 \\ \hline
    Closest Uni-Token Only        & 5.83  \\
    Multi-NMT + ULR + ($A$=$I$) & 18.61 \\ 
    Multi-NMT + ULR       & \textbf{20.01} \\ \hline
    Multi-NMT + BT & 17.91 \\
    Multi-NMT + ULR + BT & \textbf{22.35} \\  \hline
    Multi-NMT + ULR + MoLE & 20.51 \\
    Multi-NMT + ULR + MoLE + BT & \textbf{22.92} \\ \hline\hline
    Full data (612k) NMT & \textbf{28.34} \\
    \end{tabular}
    \caption{\label{ro_test1} BLEU scores evaluated on test set (6k), compared with ULR and MoLE. ``vanilla" is the standard NMT system trained only on Ro-En training set}
\end{table}

We perform  thorough experiments to examine effectiveness of the proposed method; we do ablation study on Ro-En where  all the models are trained based on the same Ro-En corpus with 6k sentences.

As shown in Table~\ref{ro_test1}, it is obvious that 6k sentences of parallel corpora  completely fails to train a vanilla  NMT model. Using Multi-NMT with the assistance of 7.8M auxiliary language sentence pairs, Ro-En translation performance gets a substantial improvement which, however, is still limited to be usable. By contrast, the proposed ULR boosts the Multi-NMT significantly with +5.07 BLEU, which is further boosted to +7.98 BLEU when incorporating sentence-level information using both MoLE and BT.  Furthermore, it is also shown that ULR works better when a trainable transformation matrix $A$ is used (4th vs 5th row in the table). Note that, although still $5\sim 6$ BLEU scores lower than the full data ($\times 100$ large) model. 

We also measure the translation quality of simply training the vanilla system while replacing each  token of the Ro sentence with its closet universal token in the projected embedding space, considering we are using the target languages (En) as the universal tokens. Although the performance is much worse than the baseline Multi-NMT, it still outperforms the vanilla model which implies the effectiveness of the embedding alignments.

\paragraph{Monolingual Data}
In Table.~\ref{ro_test1},  we also showed the performance when incorporating the monolingual Ro corpora to help the UniNMT training in both cases with and without ULR. The back-translation improves in both cases, while the  ULR  still obtains the best score  which indicates that the gains achieved are additive.

\begin{figure*}[t]
\centering
\includegraphics[width=\linewidth]{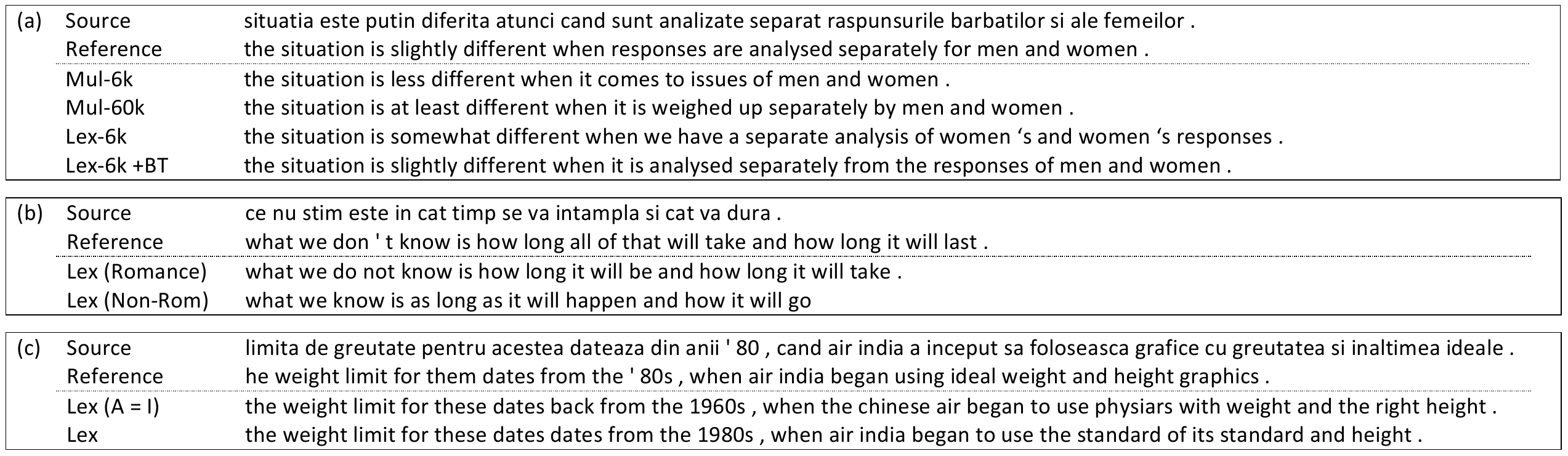} \vspace{-10pt}
\caption{\label{fig.exp}Three sets of examples on Ro-En translation with variant settings. }
\end{figure*}
\begin{figure*}[t]
\centering
\includegraphics[width=\linewidth]{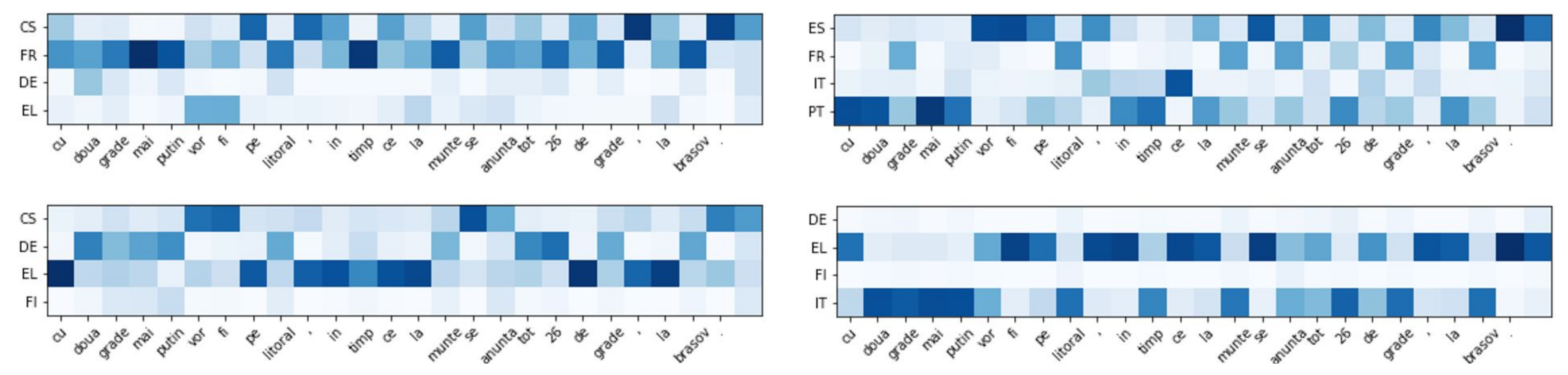}\vspace{-10pt}
\caption{\label{fig.moe} The activation visualization of mixture of language experts module on one randomly selected Ro source sentences trained together with different auxiliary languages. Darker color means higher activation score. }
\end{figure*}

\paragraph{Corpus Size}
As shown in Fig.~\ref{fig.size}, we also evaluated our methods with varied  sizes -- 0k\footnote{For 0k experiments, we used the pre-projection learned from 6k data. It is also possible to use unsupervised learned dictionary.}, 6k, 60k and 600k -- of the Ro-En corpus. The vanilla NMT and the multi-lingual NMT are used as baselines. It is clear in  all cases that the performance gets better when the training corpus is larger. However, the multilingual with ULR works much better with a small amount of training examples. Note that, the usage of ULR universal tokens also enables us to directly work on a ``pure zero" resource translation with a shared multilingual NMT model. 

\paragraph{Unknown Tokens}
One explanation on how ULR help the translation for almost zero resource languages is it greatly cancel out the effects of missing tokens that would cause out-of-vocabularies during testing. As in Fig.~\ref{fig.missing}, the translation performance heavily drops when it has more ``unknown" which cannot be found in the given 6k training set, especially for the typical multilingual NMT.  Instead, these ``unknown" tokens will naturally have their embeddings based on ULR  projected universal tokens even if we never saw them in the training set. When we apply back-translation over the monolingual data, the performance  further improves which can almost catch up with the model trained with 60k data. 





\subsection{Qualitative Analysis}
\paragraph{Examples} Figure \ref{fig.exp} shows some cherry-picked examples for Ro-En. Example (a) shows how the lexical selection get enriched when introducing ULR (Lex-6K) as well as when adding Back Translation (Lex-6K-BT). Example (b) shows the effect of using romance vs non-romance languages as the supporting languages for Ro. Example (c) shows the importance of having a trainable $A$ as have been discussed; without trainable $A$ the model confuses "india" and "china" as they may have  close representation in the mono-lingual embeddings.

\paragraph{Visualization of MoLE}
Figure \ref{fig.moe} shows the activations along with the same source sentence with various auxiliary languages. It is clear that MoLE is effectively switching between the  experts when dealing with  zero-resource language words. 
For this particular example of Ro, we can see that the system is utilizing  various auxiliary languages based on their relatedness to the source language. We can approximately rank the relatedness based of the influence of each language. For instance, the influence can be approximately ranked as $Es \approx Pt > Fr \approx It > Cs \approx El > De > Fi$, which is interestingly close to the  grammatical relatedness of Ro to these languages. On the other hand, Cs has a strong influence although it does not fall in the same language family with Ro, we think this is due to the geo-graphical influence between the two languages  since  Cs and Ro share similar phrases and expressions. This shows that MoLE learns to utilize resources from similar languages.

\subsection{Fine-tuning a Pre-trained Model}
All  the described experiments above had  the low resource languages  jointly trained  with all the auxiliary high-resource languages, where the training of the large amount of high-resource languages can be seen as a sort of regularization.  It is also common to  train a model on high-resource languages first, and then fine-tune the model on a small resource language similar to transfer learning approaches~\citep{zoph2016transfer}. However, it is not trivial to effectively fine-tune NMT models on extremely low resource data since  the models  easily over-fit due to over-parameterization of the neural networks. 

\begin{figure}
	\centering
	\includegraphics[width=\linewidth]{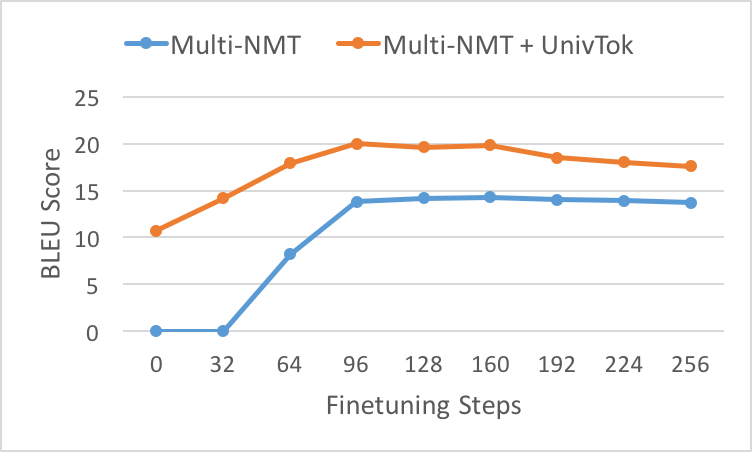}\vspace{-20pt}
	\caption{\label{fig.finetune}Performance comparison of Fine-tuning on 6K RO sentences.}

\end{figure}

In this experiment, we have explored the  fine-tuning tasks using our approach. First, we train a Multi-NMT model (with ULR)  on \{Es, Fr, It, Pt\}-En  languages only to create a zero-shot setting for Ro-En translation. Then, we start fine-tuning the model with $6k$ parallel corpora of Ro-En, with and without ULR. As shown in Fig.~\ref{fig.finetune}, both  models improve a lot over the baseline. 
With the help of ULR, we can  achieve a BLEU score of around $10.7$ (also shown in Fig.~\ref{fig.size}) for Ro-En translation with ``zero-resource" translation. The BLEU score can  further  improve to almost  $20$ BLEU after 3 epochs of training on $6k$ sentences using ULR. This is almost $6$ BLEU higher than the best score of the baseline. It is worth noting that this fine-tuning is a very efficient  process since it only takes less than 2 minutes to train for 3 epochs over such  tiny amount of data. This is very appealing  for practical applications where  adapting a per-trained system  on-line is a big advantage.  As a future work, we will further investigate a better fine-tuning strategy such as meta-learning~\citep{finn2017model} using ULR.

\section{Related Work} 
Multi-lingual NMT has been extensively studied in a number of papers such as  \newcite{lee2016fully}, \newcite{johnson2016google}, ~\newcite{zoph2016transfer} and \newcite{firat2016multi}. As we discussed, these approaches have significant limitations with zero-resource cases. \newcite{johnson2016google} is more closely related to our current approach, our work is extending  it  to overcome the  limitations with very low-resource languages and enable sharing of  lexical and sentence representation across multiple languages. 

Two recent related works are targeting the same problem of  minimally supervised  or totally unsupervised NMT. \newcite{artetxe2017unsupervised} proposed a totally unsupervised approach depending on multi-lingual embedding similar to ours and dual-learning and reconstruction techniques to train the model from mono-lingual data only. \newcite{lample2017unsupervised} also proposed a quite similar approach while utilizing adversarial learning.   

\section{Conclusion}
 In this paper, we propose a new  universal machine translation approach that enables sharing resources between high resource languages and extremely low resource languages. 
  Our approach is able to achieve 23 BLEU on Romanian-English WMT2016 using a tiny parallel corpus of 6k sentences, compared to the 18 BLEU of strong multi-lingual baseline system. 
\bibliography{naaclhlt2018}
\bibliographystyle{acl_natbib}

\end{document}